\documentclass{article}

\usepackage[final,nonatbib]{neurips_2020}

\usepackage[utf8]{inputenc} 
\usepackage[T1]{fontenc}    
\usepackage{url}            
\usepackage{cite}
\usepackage{booktabs}       
\usepackage{amsmath}
\usepackage{amsthm}
\usepackage{amsfonts}       
\usepackage{amssymb}
\usepackage{color}
\usepackage[pdftex]{graphicx}
\usepackage{subcaption}
\usepackage{wrapfig}
\usepackage{nicefrac}       
\usepackage{microtype}      
\usepackage{hyperref}       
\usepackage{algorithm}
\usepackage{algorithmic}

\usepackage{makecell}

\newcommand{\E}{\mathbb{E}}

\newcommand{\p}[2]{\frac{\partial #1}{\partial #2}}

\def\eg{{\frenchspacing\it e.g.}}
\def\etc{{\frenchspacing\it etc.}}
\def\rms{{\frenchspacing r.m.s.}}
\def\expec#1{\langle#1\rangle}
\def\tr{\hbox{tr}\,}

\def\E{\textbf{E}}

\def\V{\textbf{V}}

\def\vepsilon{\boldsymbol{\varepsilon}}

\def\ng{{n_g}}

\def\fhat{\tilde{f}}
\def\fnn{f_{\rm NN}}

\def\muhat{\widehat{\mathbf{\mu}}}
\def\p{\textbf{p}}

\def\vhat{\widehat{\bf v}}
\def\yhat{\hat{\bf y}}

\def\x{\textbf{x}}

\def\ith{i^{\rm th}}

\usepackage{amsmath}

\def\logplus{{\log_+\hskip-0.7mm}}
\def\DL{L_d}
\def\tr{\hbox{tr}\,}

\def\spose#1{\hbox to 0pt{#1\hss}}
\def\simlt{\mathrel{\spose{\lower 3pt\hbox{$\mathchar"218$}}
   \raise 2.0pt\hbox{$\mathchar"13C$}}}
\def\simgt{\mathrel{\spose{\lower 3pt\hbox{$\mathchar"218$}}
     \raise 2.0pt\hbox{$\mathchar"13E$}}}
 \def\simpropto{\mathrel{\spose{\lower 3pt\hbox{$\mathchar"218$}}
     \raise 2.0pt\hbox{$\propto$}}}

\def\beq#1{\begin{equation}\label{#1}}
\def\eeq{\end{equation}}
\def\beqa#1{\begin{eqnarray}\label{#1}}
\def\eeqa{\end{eqnarray}}
\def\eq#1{equation~(\ref{#1})}	
\def\Eq#1{Equation~(\ref{#1})}

\def\fig#1{Figure~\ref{#1}}

\def\Sec#1{Section~\ref{#1}}

\title{AI Feynman 2.0: Pareto-optimal symbolic regression exploiting graph modularity}

\author{%
Silviu-Marian Udrescu$^1$, Andrew Tan$^1$, Jiahai Feng$^1$, Orisvaldo Neto$^1$, Tailin Wu$^2$ \& Max Tegmark$^{1,3}$\\
$^1$MIT Dept.~of Physics and Institute for AI \& Fundamental Interactions, Cambridge, MA, USA\\
$^2$Stanford Dept.~of Computer Science, Palo Alto, CA, USA\\
$^3$Theiss Research, La Jolla, CA, USA\\
\texttt{$^1$\{sudrescu, aktan, fjiahai, oris,tegmark\}@mit.edu, $^2$tailin@cs.stanford.edu}
}

\begin{document}

\maketitle

\begin{abstract}
We present an improved method for symbolic regression that seeks to fit data to formulas that are Pareto-optimal, in the sense of having the best accuracy for a given complexity. It improves on the previous state-of-the-art by typically being orders of magnitude more robust toward noise and bad data, and also by discovering many formulas that stumped previous methods. We develop a method for discovering generalized symmetries (arbitrary modularity in the computational graph of a formula) from gradient properties of a neural network fit. We use normalizing flows to generalize our symbolic regression method to probability distributions from which we only have samples, and employ statistical hypothesis testing to accelerate robust brute-force search.
\end{abstract}

\section{Introduction}
\label{IntroSec}

\begin{figure}[phbt]
\centerline{\includegraphics[width=0.6\linewidth]{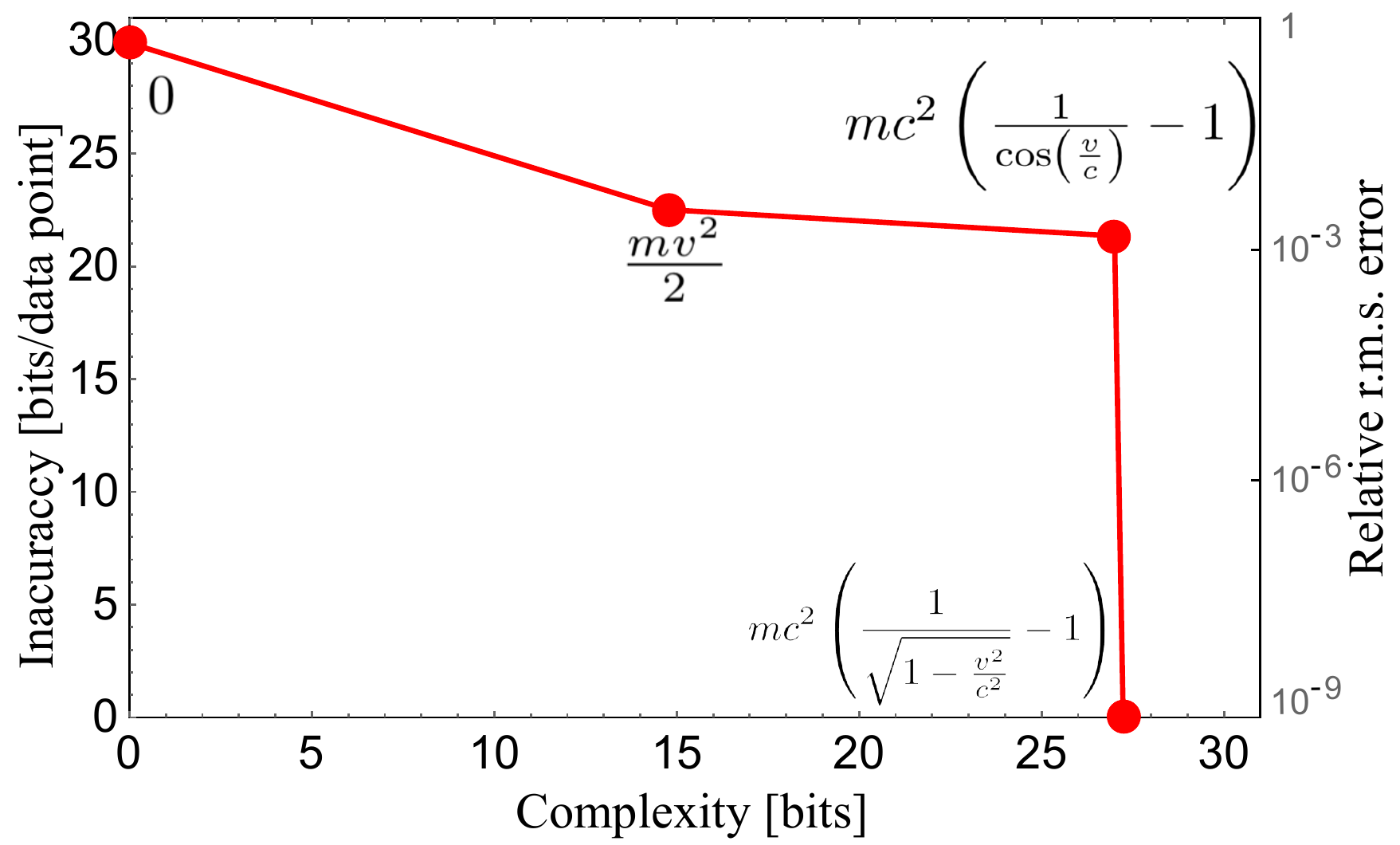}}
\caption{Our symbolic regression of data on how kinetic energy depends on mass, velocity and the speed of light discovers a Pareto-frontier of four formulas that are each the most accurate given their complexity. Convex corners reveal particularly useful formulas, in this case Einstein's formula and the classical approximation $mv^2/2$. 
\label{kineticFig}
}
\end{figure}

A central challenge in science is {\it symbolic regression}: 
discovering a symbolic expression that provides a simple yet accurate fit to a given data set.
More specifically, we are given a table of numbers, whose rows are of the form $\{x_1,...,x_n,y\}$ where
 $y=f(x_1,...,x_n)$, and our task is to discover the correct symbolic expression (composing mathematical functions from a user-provided set)
for the unknown mystery function $f$, optionally including the complication of noise and outliers.
Science aside, symbolic regression has the potential to replace some inscrutable black-box neural networks by simple yet accurate symbolic approximations, helping with the timely goal of making 
high-impact AI systems more interpretable and reliable \cite{russell2015research, amodei2016concrete, boden2017principles, krakovna2016increasing,russell2019human}.

Symbolic regression is difficult because of the exponentially large combinatorial space of symbolic expressions.
Traditionally, it has relied on human intuition, leading to the discovery of some of the most famous formulas in science. 
More recently, there has been great progress toward fully automating the process
\cite{crutchfield1987equation, dzeroski1995discovering, bradley2001reasoning, langley2003robust, schmidt2009distilling, mcree2010symbolic, searson2010gptips, dubvcakova2011eureqa, stijven2011separating, schmidt2011automated, hillar2012comment, daniels2015automated, langley2015heuristic, arnaldo2015building, brunton2016discovering, guzdial2017game, quade2018sparse, koch2018mutual, kong2018new, udrescu2020ai, liang2019phillips}, and open-source software now exists that can discover quite complex physics equations by combining neural networks with techniques inspired by physics and information theory \cite{udrescu2020ai}. 
Although \cite{udrescu2020ai} achieved state-of-the-art performance using a neural network approximation of the unknown function to discover simplifying function properties, it did so in an unprincipled and ad hoc way that we replace by a general, principled and much more effective method, incorporating four main contributions:
\begin{enumerate}
\item We recursively exploit modularity in the function's computational graph.
While  \cite{udrescu2020ai} discovered merely two types of graph modularity (symmetry and separability) involving merely four particular bivariate functions ($+$, $-$, $\times$ and $\div$), our method has the potential to discover {\it any} graph modularity involving {\it any} functions of $n=2,3,...$ variables, by examining gradients of the neural network fit.
\item Instead of concluding that a candidate function or graph decomposition is good because the fitting accuracy exceeds arbitrary hyperparameter-determined thresholds,  we eliminate these hyperparameters and use a Pareto-frontier 
(of description-length complexity versus inaccuracy)
to prune our search over candidate expressions by discarding all candidates not on the frontier, 
improving robustness to noise and bad data. 
\item Instead of simply rejecting formula candidates using $L_\infty$-norm (rejecting as soon as the error for a single data point crosses a threshold), we reject using statistical hypothesis testing, further improving robustness.
\item We use normalizing flows to enable 
symbolic 
regression of probability distributions from samples.
\end{enumerate}
This enables more complex formulas to be discovered and improving noise robustness by 1-3 orders of magnitude.
We describe our symbolic regression algorithm (which is publicly available\footnote{Our code is can be installed by typing {\it pip install aifeynman} and is also available at \url{https://ai-feynman.readthedocs.io}.}) 
in \Sec{MethodsSec} and test it with numerical experiments in \Sec{ResultsSec}.

\section{Method}
\label{MethodsSec}

Our symbolic regression algorithm uses a divide-and-conquer approach as in \cite{udrescu2020ai}. 
We directly solve a mystery 
in two base cases: if the mystery function 
$f(x_1,...,x_n)$ is a low-order polynomial or if it is simple enough to be discovered by brute-force search.
Otherwise, we recursively try the 
strategies that we will now describe for
replacing it by one or more simpler mysteries, ideally with fewer input variables.

\subsection{Leveraging graph modularity against the curse of dimensionality}
\label{ModularitySubsec}

\begin{figure}[phbt]
\centerline{\includegraphics[width=\linewidth]{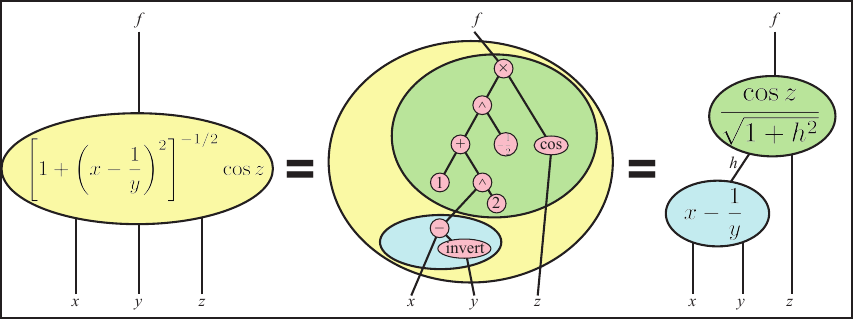}}
\caption{All functions can be represented as tree graphs whose nodes represent a set of 
basic functions (middle panel). Using a neural network trained to fit a mystery function (left panel),
our algorithm seeks a decomposition of this function into others with fewer input variables (right panel), in this case of the form $f(x,y,z)=g[h(x,y),z]$ .
\label{modulesFig}
}
\end{figure}

When we define and evaluate a mathematical function, we typically represent it as composed of some basis set $S$ of simpler functions. As illustrated in \fig{modulesFig} (middle panel), this representation can be specified as a graph whose nodes contain elements of $S$. The most popular basis functions in the scientific literature tend to
be functions of two variables (such as $+$ or $\times$), one variable (such as $\sin$ or $\log$) or no variables (constants such as $2$ or $\pi$).
For many functions of scientific interest, this graph is {\it modular} in the sense that it can be partitioned in terms of functions with fewer input variables, as in \fig{modulesFig} (right panel).

A key strategy of our symbolic regression algorithm is to recursively discover such modularity, thereby reverse-engineering the computational graph of a mystery function, starting with no information about it other than an input-output data table.
This is useful because there are exponentially many ways to combine $n$  basis functions into a module, making it extremely slow and difficult for brute-force or genetic algorithms to discover the correct function when $n$ is large. Our divide-and-conquer approach of first breaking the function into smaller modules with smaller $n$ that can be solved separately thus greatly accelerates the solution.
We implement this modularity discovery algorithm in two steps:
\begin{enumerate}
\item Use the user-provided data table  
 to train a neural network $\fnn(\x)$ that accurately approximates the mystery function $f(\x)$.
\item Perform numerical experiments on $\fnn(\x)$ to discover graph modularity. 
\end{enumerate}

\begin{figure}[bt]
\centerline{\includegraphics[width=\linewidth]{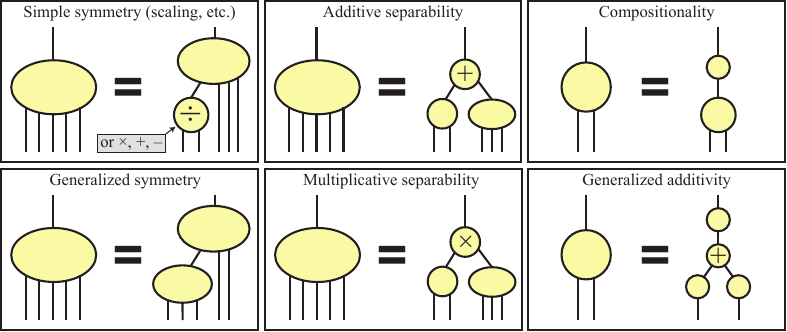}}
\caption{Examples of graph modularity that our algorithm can auto-discover. 
Lines denote real-valued variables and ovals denote functions, with larger ones being more complex.
\label{modularityFig}
}
\end{figure}

Specifically, we test for the six types of graph modularity illustrated in \fig{modularityFig} and listed in Table~\ref{strategyTable}, and choose between the discovered candidates as described in \Sec{ParetoSubsec}.
Our method for discovering separability is described in \cite{udrescu2020ai}.
As we will see below, all our other types of graph modularity 
(compositionality, symmetry and generalized additivity)
can be revealed by $\nabla f$, the gradient of our mystery function $f$.

\begin{table}[tb]
  \caption{Simplification strategies}
  \label{strategyTable}
  \centering
  \begin{tabular}{lll}
\toprule
Name   			& Property&Action\\
\toprule
Negativity 			&$f(x_1,x_2,...)<0$					&Solve for $g\equiv -f$\\
\midrule
Positivity 			&$f(x_1,x_2,...)>0$					&Solve for $g\equiv\ln f$\\
\midrule
Additive 			&$f(x_1,...,x_k,x_{k+1},...,x_n)=$		&\\
separability				&$g(x_1,...,x_k)+h(x_{k+1},...,x_n)$		&Solve for $g$ \& $h$\\
\midrule
Multiplicative 		&$f(x_1,...,x_k,x_{k+1},...,x_n)=$		&\\
separability				&$g(x_1,...,x_k)h(x_{k+1},...,x_n)$		&Solve for $g$ \& $h$\\
\midrule
Simple symmetry	&$f(x_1,x_2,...)=g(x_1\odot x_2,...),\>\>\>\odot\in\{+,-,\times,/\}$			&Solve for $g$\\
\midrule
Compositionality 		&$f(x_1,...,x_n)=g(h(x_1,...,x_n)),\>\>\>\hbox{$h$ simpler than $f$}$	&Find $h$ with $\nabla h\propto\nabla f$\\
\midrule
Generalized 			&$f(x_1,...,x_k,x_{k+1},...,x_n)=$		&Find $h$ satisfying \\
symmetry				&$g[h(x_1,...,x_k),x_{k+1},...,x_n]$		&${\partial h\over\partial x_i}\propto {\partial f\over\partial x_i}$, $i=1,...,k$\\
\midrule
Generalized 			&$f(x_1,x_2)=F[g(x_1)+h(x_2)]$	&Solve for $F$, $g$ \& $h$\\
additivity				&							&\\
\midrule
Zero-snap				&$\fhat$ has numerical parameters	$\p$&Replace $p_i$ by $0$\\
\midrule
Integer snap			&$\fhat$ has numerical parameters	$\p$&Round $p_i$ to integer\\
\midrule
Rational snap			&$\fhat$ has numerical parameters	$\p$&Round $p_i$ to fraction\\
\midrule
Reoptimize			&$\fhat$ has numerical parameters	$\p$&Reoptimize $\p$ to\\
	 				&						       &minimize inaccuracy\\
\bottomrule
\end{tabular}
\end{table}

\paragraph{Compositionality}

Let us first consider the case of {\it compositionality} (\fig{modularityFig}, top right), where 
$f(\x)=g(h(\x))$ and $h$ is a scalar function simpler than $f$ in the sense of being expressible with a smaller graph as in \fig{modularityFig}. 
By the chain rule, we have 
\beq{CompositionalityCriterionEq}
\nabla f(\x) = g'(h(\x))\nabla h(\x),\quad\hbox{so}\quad
\widehat{\nabla f}=\pm\widehat{\nabla h},
\eeq
where hats denote unit vectors: 
$\widehat{\nabla f}\equiv\nabla f/|\nabla f|$, {\etc}
This means that if we can discover a function $h$ whose gradient is proportional to that of $f$ (which we will describe a process for in \Sec{bruteSec}), then we can simply replace the variables $\x$ in the original mystery data table by the single variable $h(\x)$ and recursively apply our AI Feynman algorithm to the new one-dimensional symbolic regression problem of discovering $g(h)$.

\paragraph{Generalized symmetry}

Let us now turn to {\it generalized symmetry} (\fig{modularityFig}, bottom left), where 
$k$ of the $n$ arguments enter only via some scalar function $h$ of them. 
Specifically, we say that an $f$ has generalized symmetry if
the $n$ components of the
vector $\x\in\mathbb{R}^n$ can be split into groups of $k$ and $n-k$ components (which we denote  by the vectors $\x'\in\mathbb{R}^k$  and $\x''\in\mathbb{R}^{n-k}$) 
such that $f(\x)=f(\x',\x'')=g[h(\x'),\x'']$ for some function $g$.
By the chain rule, we have 
\beq{GenSymCriterionEq}
\nabla_{\x'} f(\x',\x'') = g_1[h(\x'),\x'']\nabla h(\x'),\quad\hbox{so}\quad
\widehat{\nabla_{\x'} f}=\pm\widehat{\nabla h},
\eeq
where $g_1$ denotes the derivative of $g$ with respect to its first argument.
This means that $\widehat{\nabla_{\x'}}f(\x',\x'')$ is independent of $\x''$, which it would not be for a generic function $f$.
$\x''$-independence of the normalized gradients  $\vhat(\x',\x'')\equiv\widehat{\nabla_{\x'}}f(\x',\x'')$ thus provides a 
smoking gun signature of generalized symmetry.
Whereas our compositionality discovery above requires discovering an explicit function $h$, we can discover generalized symmetry without knowing $h$, thus only performing the time-consuming task of searching for an $h$ satisfying \eq{GenSymCriterionEq} after determining that a solution exists. 
The Supplementary Material details how we numerically test for $\x''$-independence of $\vhat(\x',\x'')$.

\paragraph{Generalized additivity}

If $f$ is a function of two variables, then we also test for {\it generalized additivity}
(\fig{modularityFig}, bottom right), where 
$f(x_1,x_2) = F[g(x_1)+h(x_2)]$.
If we define the function
\beq{GenAddCriterionEq}
s(x_1,x_2)\equiv {\partial f/\partial x_1\over \partial f/\partial x_2},\quad\hbox{then}\quad
s(x_1,x_2)={g'(x_1)\over h'(x_2)}
\quad
\eeq
if $f$ satisfies the generalized additivity property.
In other words, we simply need to test if $s$ is of the multiplicatively separable form 
$s(x_1,x_2)=a(x_1)b(x_2)$, and we do this using a variant of the separability test described in \cite{udrescu2020ai}. 
The Supplementary Material details how we perform this separability test numerically.

\subsection{Robustness through recursive Pareto-optimality}
\label{ParetoSubsec}

As illustrated in \fig{kineticFig}, the goal of our symbolic regression of a data set is to approximate $f(\x)$ by functions $\fhat(\x)$ that are not only accurate, but also simple, in the spirit of Occam's razor. 
As in \cite{schmidt2009distilling}, we seek functions that are {\it Pareto-optimal} in the sense of there being no other function that is both simpler and more accurate.
We will adopt an information-theoretical approach and use bits of information to measure lack of both accuracy and simplicity. 
 
\begin{figure}[bt]
\centerline{\includegraphics[width=\linewidth]{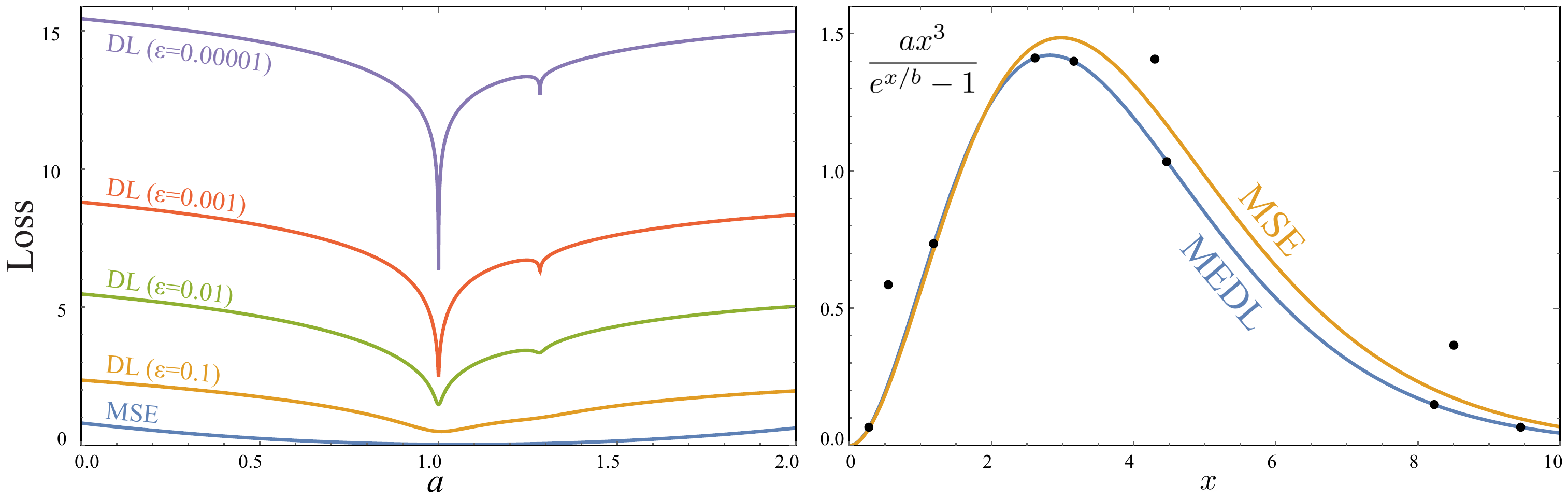}}
\caption{When fitting a function (the right panel shows the example ${ax^3\over e^{x/b}-1}$) to data with outliers, 
minimizing mean-squared-error (MSE) biases the curve toward the outliers (here finding $a\approx 0.89$, $b\approx 1.056$), whereas minimizing mean error description length (MEDL) ignores the outliers and recovers the correct answer $a=b=1$. Left panel compares MSE and MEDL loss functions for $b=1$. 
\label{mdlFig}
}
\end{figure}

For {\it accuracy}, we wish the vector $\vepsilon$ of prediction errors $\varepsilon_i\equiv y_i-\fhat(\x_i)$ to be small. We quantify this not by the 
mean-squared error $\expec{\varepsilon_i^2}$ 
or max-error $\max |\varepsilon_i|$ as in \cite{schmidt2009distilling,udrescu2020ai},
but by the MEDL, the {\it mean error-description-length} $\expec{\DL(\varepsilon_i)}$
defined in Table~\ref{complexityTable}.
As argued in \cite{wu2019toward} and illustrated in \fig{mdlFig}, this improves robustness to outliers.
We analogously quantify {\it complexity} by the description length $\DL$ defined as in 
\cite{wu2019toward}, summarized in Table~\ref{complexityTable}.

$\DL$ can be viewed as a crude but computationally convenient approximation of 
the number of bits needed to describe each object, made differentiable where possible. We choose the precision floor $\epsilon\equiv 2^{-30}\sim 10^{-9}$.
For function complexity, both input variables and mathematical functions (\eg, $\cos$ and +) count toward $n$ and $k$.
For example, the classical kinetic energy formula has
$\DL(``m\times v\times v/2'') = \DL(2) + k\log_2 n=
\log_2 3 + 6\log_2 4 \approx 13.6$ bits, since the formula contains
$n=4$ basis functions ($m$, $v$, $\times$ and $/$) used 
$k=6$ times.

\begin{table}[tb]
  \caption{Complexity definitions}
  \label{complexityTable}
  \centering
  \begin{tabular}{lll}
    \toprule
Object   			& Symbol	&Description length $\DL$\\
\midrule
Natural number		&$n$		&$\log_2 n$\\
Integer 			&$m$	&$\log_2 (1+|m|)$\\
Rational number	&$m/n$	&$\DL(m)+\DL(n)=\log_2[(1+|m|)n]$\\
Real number		&$r$		&$\logplus\left({r\over\varepsilon}\right),\quad\logplus(x)\equiv{1\over 2}\log_2\left(1+x^2\right)$\\
Parameter vector	&$\p$	&$\sum_i\DL(p_i)$\\
Parametrized function&$f(\x;\p)$&$\DL(\p)+k\log_2 n$; $n$ basis functions appear k times\\
\bottomrule
\end{tabular}
\end{table}

We wish to make the symbolic regression implementation of \cite{udrescu2020ai} more robust; it sometimes fails to discover the correct expression because of noise in the data or inaccuracies introduced by the neural network fitting. The neural network accuracy may vary strongly with $\x$, becoming quite poor in domains with little training data or when the network is forced to extrapolate rather than interpolate, and we desire a regression method robust to such outliers.
We expect our insistence on Pareto-optimal functions in the information plane of \fig{kineticFig} to increase robustness, both because $\expec{\DL(\varepsilon_i)}$ is robust (\fig{mdlFig})
and because noise and systematic errors are unlikely to be predictable by a simple mathematical formula with small $\DL$. More broadly, minimization of total exact description length (which $\DL$ crudely approximates) provably avoids the overfitting problem that plagues many alternative machine-learning strategies \cite{rissanen1978modeling,hutter2000theory,grunwald2005advances}. 

\paragraph{Speedup by recursive Pareto frontier composition}

When recursively symbolically regressing various modules (see \fig{modulesFig}), we end up with a Pareto frontier of candidate functions for each one.
If there are $n_i$ functions on the $\ith$ frontier, 
then combining them all would produce $\prod_i n_i$ candidates $\fhat(\x)$ for the original function 
$f(\x)$. We speed up our algorithm by Pareto-pruning after each merge step: whenever two modules are combined (via composition or multiplication, say), the resulting $n_1 n_2$ functions are pruned by removing all functions that are Pareto-dominated by another function that is both simpler and more accurate. 
Pruning models on the Pareto frontier significantly reduces the number of models that need to be evaluated, since in typical scenarios, the number of Pareto-optimal points grows only logarithmically with the total number of points.

\paragraph{Robust speedup of brute-force graph search with hypothesis testing}
\label{bruteSec}

Our recursive reduction of regression mysteries into simpler ones terminates at the base case when 
the mystery function has only one variable and cannot be further modularized. 
As in \cite{udrescu2020ai}, we subject these (and also all multivariate modules) to two solution strategies, polynomial fitting up to some degree ($4$ by default) and brute-force search, and then add all candidates functions to the Pareto plane and prune as above.
The brute-force search would, if run forever, try all symbolic expressions by looping over ever-more-complex graphs  (the middle panel of \fig{modulesFig} shows an example) and over function options for each node.  

Our brute-force computation of the 
Pareto frontier simply tries all functions $f_k(\x)$ ($k=1,2...$) in order of increasing complexity $\DL(f_k)$ and keeps only those with lower mean error-description-length $d_k\equiv{1\over N}\sum_{i=1}^N d_{ki}$ than the previous record holder, where 
$d_{ki}\equiv\DL[y_i-f_k(\x_i)]$.
When instead fitting normalized gradient vectors $\widehat{\nabla f}$ as in \Sec{ModularitySubsec}, 
we define $d_{ki}\equiv\DL[1-|\yhat_i\cdot\widehat{\nabla f_k}(\x_i)|]$ to handle the sign ambiguity. The bad news is that computing $d_k$ exactly is slow, requiring evaluation of $f_k(\x_i)$ for all $N$ data points $\x_i$.
The good news is that this is usually unnecessary, since for the vast majority of all candidate functions, it becomes obvious that they provide a poor fit after trying merely a handful of data points. We therefore accelerate the search via the following procedure.
Before starting the loop over candidate functions, we sort the data points in random order to be able to interpret 
the numbers $d_{ki}$ as random samples from a probability distribution whose mean is the sought-for $d_k$ and whose standard deviation is $\sigma_k$. Let $d_{k*}$ and $\sigma_{k*}$ denote the corresponding quantities that were computed for the 
previous best-fit function we added to the Pareto frontier.
We make the simplifying approximations that $\sigma_k=\sigma_{k*}$ and that all errors are uncorrelated, so that the loss
estimate from the first $m$ data points
$\bar{d}_{km}\equiv {1\over m}\sum_{i=1}^m d_{ki}$ has mean $d_k$ and standard deviation $\sigma_{k*}/\sqrt{m}$.
We now test our candidate function $f_i$ on one data point at a time and reject it as soon as 
\beq{RejectionEq}
z>\nu,\quad\hbox{where}\quad z\equiv\sqrt{m}\>{\bar{d}_{km}-d_{k*}\over\sigma_{k*}},
\eeq
where $\nu$ is a hyperparameter that we can interpret as the ``number of sigmas" we require to rule out a candidate function as viable 
when its average error exceeds the previous record holder. 
We find that $\nu=10$ usually works well, generically requiring no more than a handful of evaluations $m$ per candidate function asymptotically.
We can further increase robustness by increasing $\nu$ at the price of longer runtime.

\paragraph{Speedup by greedy search of simplification options}

We do not {\it a priori} know which of the modular decompositions from \fig{modularityFig} are most promising, and recursively trying all combinations of them would involve trying exponentially many options. 
We therefore accelerate our algorithm with a greedy strategy where at each step we compare the decomposition in a unified way and try only the most accurate one --- our runtime thus grows roughly linearly with $n$, the number of input variables.
$f(\x)$ stays constant along constant-$h$ curves for
generalized symmetry, simple symmetry (where $h(x,y)=x+y$, $x-y$, $xy$ or $x/y$) and 
generalized additivity (where $h(x,y)=a(x)+b(y)$).
We thus test the accuracy of all such $h$-candidates by starting at a datapoint $\x_i$ and computing an error $\epsilon_i\equiv f(\tilde{\x}_i)-f(\x_i)$ for some $\tilde{\x}_i$
satisfying $h(\tilde{\x}_i)=h(\x_i)$. For additive and multiplicative separability, we follow \cite{udrescu2020ai} by examining a rectangle in parameter space and predicting $f$ at the fourth corner from the other three, defining $\epsilon_i$ as the mismatch. The supplementary material details how our test points are chosen.

After this greedy recursive process has terminated, we further improve the Pareto frontier in two ways. We first add models where rational numbers are replaced by reals and optimized by gradient descent to fit the data. We then add models with zero-snap, integer-snap and rational-snap from Table~\ref{strategyTable} applied to all real-valued parameters as described in \cite{wu2019toward}, pruning all Pareto-dominated models after each step. For example, if there are 3 real-valued parameters, integer-snap generates 3 new models where the 1, 2 and 3 parameters closest to integers get rounded, respectively.

\subsection{Leveraging normalizing flows to symbolic regress probability distributions}
\label{FlowSubsec}

\begin{figure}[bt]
\centerline{\includegraphics[width=\linewidth]{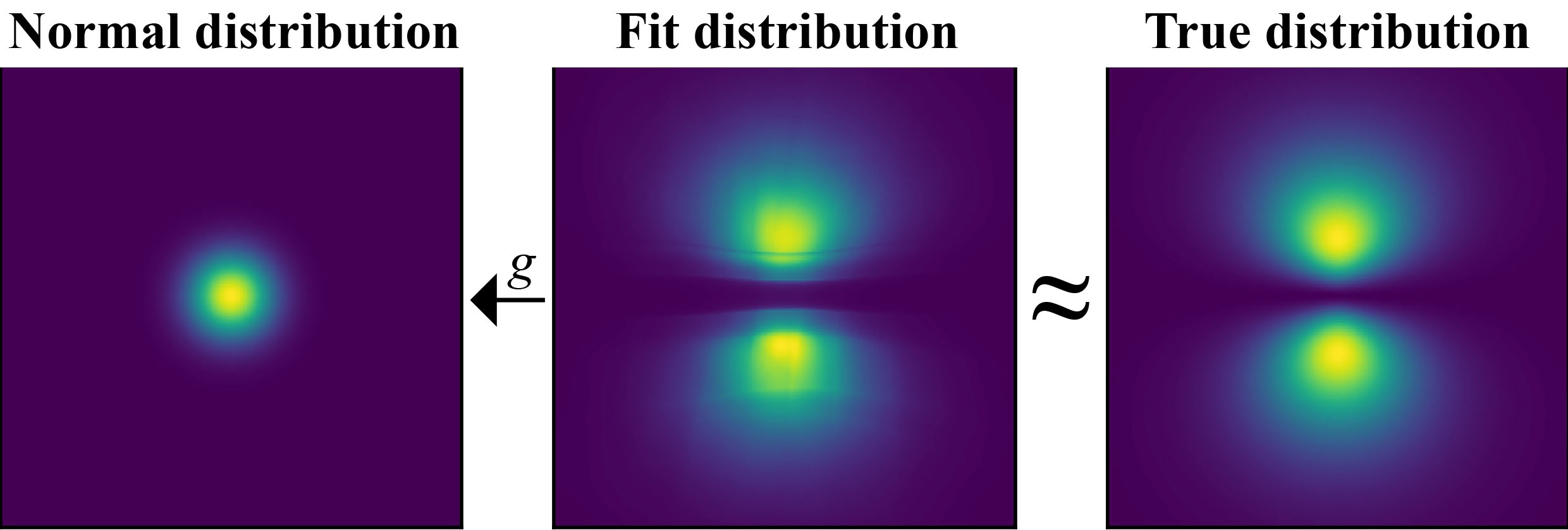}}
\caption{
\label{nfdemo}
A normalizing flow $g$ maps samples from a probability distribution $f$ (right) into a normal distribution (left), enabling an estimate (middle) of $f$, here illustrated for the $n=2$, $l=1$, $m=0$ hydrogen orbital from Table~\protect\ref{ProbDistTable}.
}
\end{figure}

An important but more difficult symbolic regression problem is when the unknown function $f(\x)$ is a probability distribution from 
which we have random samples $\x_i$ rather than direct evaluations $y_i=f(\x_i)$.
We tackle this by adding preceding the regression by a step that estimates $f(\x)$.
For this step, we use the popular {\it normalizing flow} technique \cite{rezende15,dinh2017,kingma2018,durkan2019,kobyzev2020},
training an invertible neural network mapping $\x\mapsto\x'\equiv g(\x)$ such that $\x'$ has a multivariate normal distribution $n(\x')$
as illustrated in \fig{nfdemo}. We then obtain our estimator $\fnn(\x)=n[g(\x)]|J|$, where $J$ is the Jacobian of $g$. 

We find rational-quadratic neural spline flows (RQ-NSF) suitable for relatively low-dimensional applications due to their enhanced expressivity.
Specifically, we used three steps of the RQ-NSF with RQ-NSF (C) coupling layers as described in \cite{durkan2019}, 
parametrized by three 16-neuron softplus layers, trained for $50,000$ epochs with the Adam optimizer. The learning rate was initialized to $3 \times 10^{-4}$ and halved every time the test loss failed to improve for $2500$ epochs. 

\subsection{Neural Network training}
\label{NNtrain}

Our neural network approximation $\fnn$ of the mystery function $f$ is fully-connected, feed-forward neural network with 4 hidden layers
of 128, 128, 64 and 64 neurons, respectively, all with \textit{tanh} activation function. We used 80$\%$ of the available data the training and the rest for validation. We used the $\rms$ error loss function and the Adam optimizer with $\beta$-parameters of $0.9$ and $0.999$. The learning rate was initialized to $0.01$ and reduced by a factor of 10 whenever the validation loss failed to improve for more than 20 epoch, until it reached $10^{-5}$. 
As mentioned, all our code is available at \url{https://ai-feynman.readthedocs.io} and by typing ``pip install aifeynman".

\section{Results}
\label{ResultsSec}

We now turn to quantifying the performance of our method with numerical experiments, comparing it with that of \cite{udrescu2020ai} which recently exceded the previous state of-the-art performance of \cite{schmidt2009distilling}.
To quantify robustness to noise, we add Gaussian noise of standard deviation $10^{r}$ to $y_i$ and determine the largest integer $r<0$ for which the method successfully discovers the correct mystery function $f(\x)$.
As seen in Table~\ref{robustnessTable}, our method 
solves 73 of 100 baseline problems from the
Feynman Symbolic Regression Database \cite{udrescu2020ai}
with $r=-1$, and is typically 1-3 orders of magnitude more robust than that 
of \cite{udrescu2020ai}. 
Crudely speaking, we found that adding progressively more noise shifted the most accurate formula straight upward in the Pareto plane (\fig{kineticFig}) 
until it no longer provided any accuracy gains compared with simpler approximations.

\begin{table}
  \caption{Robustness to noise}
  \label{robustnessTable}
\centerline{\includegraphics[width=0.6\linewidth]{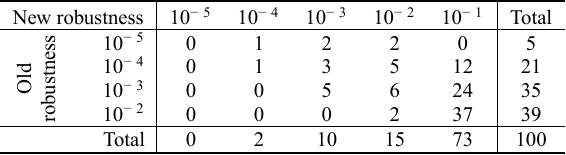}}
\end{table}

To quantify the ability of our method to discover more complex equations, 
we reran it on all 17 mysteries that \cite{udrescu2020ai} tackled and failed to solve.
We also tested a dozen new mysteries exhibiting various forms of graph modularity (see Table~\ref{MysteryTable}) that were all chosen before any of them were tested. 
Allowing at most two hours of runtime, 
the method of \cite{udrescu2020ai} solved 5 of the equations, 
whereas our new method solved them all, 
as well as four of the outstanding mysteries from \cite{udrescu2020ai} (rows 1-4).
For these first four, our method got the numerical parameters in the right ballpark with rational approximations, then discovered their exact values through gradient descent. 

\begin{table*}[tb]
\centering
\caption{\label{MysteryTable}
Test equations exhibiting 
translational symmetry $h=x\pm y$ (T), 
scaling symmetry $h=x/y$ (S),
product symmetry $h=xy$ (P),
generalized symmetry (G), 
multiplicative separability (M),
compositionality (C) and
generalized additivity (A).
}
\begin{tabular}{r|l|c}
\toprule
&Equation   & Symmetries\\
\midrule
1&$\delta = -5.41+4.9\frac{\alpha-\beta+\gamma/\chi}{3\chi}$ &TC\\
2&$\chi = 0.23+14.2\frac{\alpha+\beta}{3\gamma}$ &TS\\
3&$\beta = 213.80940889\left(1-e^{-0.54723748542\alpha}\right)$ &\\
4&$\delta = 6.87 + 11\sqrt{\alpha\beta\gamma}$ &P\\
5&$V=\left[R_1^{-1}+R_2^{-1}+R_3^{-1}+R_4^{-1}\right]^{-1}I_0\cos\omega t$\quad\hbox{(Parallel resistors)}&PGSM\\
6&$I_0={V_0\over\sqrt{R^2+\left(\omega L-{1\over\omega C}\right)^2}}$\quad\hbox{(RLC circuit )}&MG\\
7&$I={V_0\cos\omega t\over\sqrt{R^2+\left(\omega L-{1\over\omega C}\right)^2}}$\quad\hbox{(RLC circuit)}&MG\\
8&$V_2 = (\frac{R_2}{R_1+R_2}-\frac{R_x}{R_x+R_3})V_1$\quad\hbox{(Wheatstone bridge)} &SGMA\\
9&$v = c\frac{(v_1 + v_2 + v_3)/c + v_1v_2v_3/c^3}{1 + (v_1v_2 + v_1v_3 + v_2v_3)/c^2}$\quad\hbox{(Velocity addition)} &AG\\
10&$v = c \frac{(v_1 + v_2 + v_3 + v_4)/c + (v_2 v_3 v_4 + v_1 v_3 v_4 + v_1 v_2 v_4 + v_1 v_2 v_3)/c^3}{1 + (v_1 v_2+ v_1 v_3+ v_1 v_4+ v_2 v_3+ v_2 v_4+ v_3 v_4)/c^2 + v_1 v_2 v_3 v_4/c^4}$\quad\hbox{(Velocity addition)}&GA\\
11&$z = (x^4+y^4)^{1/4}$\quad\hbox{($L_4$-norm)}&AC\\
12&$w = xyz-z\sqrt{1-x^2}\sqrt{1-y^2}-y\sqrt{1-x^2}\sqrt{1-z^2}-x\sqrt{1-y^2}\sqrt{1-z^2}$ &GA\\
13&$z = \frac{xy+\sqrt{1-x^2-y^2+x^2y^2}}{y\sqrt{1-x^2}-x\sqrt{1-y^2}}$ &A\\
14&$z = y\sqrt{1-x^2}+x\sqrt{1-y^2}$ &A\\
15&$z = xy-\sqrt{1-x^2}\sqrt{1-y^2}$ &A\\
16&$r = \frac{a}{\cot{(\alpha/2)}+\cot{(\beta/2)}}$\quad\hbox{(Incircle)}&GMAC\\
\bottomrule
\end{tabular}
\vskip-5mm
\end{table*}

\begin{table*}[tb]
\centering
\caption{Probability distributions and number of samples $N$ required to discover them
\label{ProbDistTable}
}
\begin{tabular}{l|l|c}
\toprule
Distribution Name & Probability distribution & $N$\\
\midrule       
Laplace distribution & $\frac{1}{2}e^{-|x|}$ & $10^2$\\
Beta distribution ($\alpha = 0.5$, $\beta = 0.5$) & $\frac{1}{\pi}\frac{1}{\sqrt{x(1-x)}}$ & $10^4$\\
Beta distribution ($\alpha = 5$, $\beta = 2$) & $30x^4(1-x)$ & $10^4$\\
Harmonic oscillator ($n=2$, $\frac{m\omega}{\hbar} = 1$) & $\frac{2}{\sqrt\pi}x^2e^{-x^2}$ & $10^5$\\
Sinc diffraction pattern & $\frac{1}{\pi}\left(\frac{\sin{x}}{x}\right)^2$ & $10^4$\\
2D normal distribution (correlated) & $\frac{1}{\sqrt{3}\pi}e^{-\frac{2}{3}(x^2-xy+y^2)}$ & $10^3$\\
2D harmonic oscillator ($n=2$, $m=1$, $\frac{m\omega}{\hbar} = 1$) & $\frac{2}{\pi}x^2e^{-x^2-y^2}$ &$10^5$\\
Hydrogen orbital ($n=1$, $l=0$, $m=0$) & $\frac{1}{\pi}e^{-2r}$ & $10^3$\\
Hydrogen orbital ($n=2$, $l=1$, $m=0$) & $\frac{1}{16}r^2e^{-r}\cos^2\theta$ & -\\
Hydrogen orbital ($n=3$, $l=1$, $m=0$) & $\frac{1}{729}r^2\left(4-\frac{2r}{3}\right)^2e^{-\frac{2r}{3}}\cos^2\theta$ & -\\
\bottomrule
\end{tabular}
\vskip-2mm
\end{table*}

To quantify the ability of our method to discover probability distributions, 
we tested it on samples from the ten distributions in Table~\ref{ProbDistTable}. As seen in the table, 80\% were solved, requiring between $10^2$ and $10^5$ samples $\x_i$.
The flows trained in about 20 minutes on one CPU, scaling roughly linearly with sample size and number of network weights.

We discuss common failure modes below. Interestingly, these do not include overfitting, for multiple reasons:
(1)~We early-stop training when the validation loss starts increasing.  
(2)~We avoid using our neural network (to guess symbolic functions) outside the domain where it was trained.
(3)~Overfitting noise would generally {\it reduce} apparent graph modularity, thus causing failure to discover formulas rather than discovery of spurious ``overfit'' formulas.
(4)~A key desirable feature of the minimum-description-length formalism (the information-theoretical inspiration for our method) is that it provably avoids overfitting as shown in \cite{rissanen1978modeling,grunwald2005advances}.

\section{Conclusions}
\label{ConclusionsSec}

We have presented a symbolic regression method that exploits 
neural networks, graph modularity, hypothesis testing and normalizing flows, and released it at \url{https://ai-feynman.readthedocs.io}.
It improves state-of-the-art performance both by being more robust towards noise and by solving harder problems, including symbolic density estimation.

Despite these advances, numerous equations remained unsolved, motivating further work.
Here are some interesting failure modes we identified.
Some cases failed because the form of the equation precluded our method from breaking it into small enough pieces. For example, for the equation 
$$\alpha^3  e^{-\alpha}\cos\alpha\sin\alpha\left[\cos\alpha(\sin\alpha)^2-1\right](\beta-5),$$ 
our algorithm discovered the multiplicative separability into terms including only $\alpha$ and only $\beta$. However, the remaining $\alpha$-term was too complicated to be solved in a reasonable amount of time by the brute force code, and none of the graph modularity methods apply because 
they only help for functions of more than one variable.
For other equations, our method fails but not irreparably. For example, for  the function
$$22 - 4.2\left[\cos\alpha-\tan\beta\right]\tanh\gamma/\sin\chi,$$ 
our code is able to discover that $\gamma$ and $\chi$ can be separated from the rest of the equation. However, given that we allow the brute force code to run for only a minute for each iteration, the expression $\tanh\gamma$ is not discovered, mainly because we did not include $\tanh$ as one of the functions used, so the brute force would have to write that as $\frac{e^{2x}-1}{e^{2x}+1}$. 
By allowing the code to run for longer or using with other basis functions (such as $\tanh$), the code could solve this and several other mysteries that we reported as failures.
Success and failure modes are further discussed in the Supplementary Material.

There are many obvious ways in which the core ideas of this paper can be extended to 
further improve symbolic regression. For example, gradients can reveal more types of graph modularity than the \fig{modularityFig} examples that we exploited (e.g. where modules output more than one variable), additional simplification strategies can be included in the Pareto-optimal recursion, and flow-based regression can be used for regularized 
density estimation from sparse high-dimensional data. Larger and more challenging collections of science-based equations are needed to benchmark and inspire improved algorithms. 
A higher-level direction for improvement is to generalize the problem itself: Whereas symbolic regression takes the regression variables as given, discovering novel formulas from real-world data typically requires also the pre-regression step of mapping high-dimensional data into
a low-dimensional latent space whose coordinates are promising candidates for symbolic regression; \cite{udrescu2020symbolic} provides a literature review and early steps in this direction.

Pareto-optimal symbolic regression has the power to not only discover exact formulas, but also 
approximate ones that are useful for being both accurate and simple.
The mainstream view is that all known science formulas are such approximations. 
We live in a golden age of research with ever-larger datasets produced by both experiments and numerical computations, and we look forward to a future when symbolic regression is as ubiquitous as linear regression is today, 
helping us better understand the relations hidden in these datasets.

\clearpage
 
\begin{ack}
The authors with to thank Philip Tegmark for helpful comments,  and the Center for Brains, Minds, and Machines (CBMM) for hospitality.

{\bf Funding:} This work was supported by The Casey and Family Foundation, the Ethics and Governance of AI Fund, the Foundational Questions Institute, the Rothberg Family Fund for Cognitive Science and the Templeton World Charity Foundation, Inc. 

{\bf Competing interests:} The authors declare that they have no competing interests. 

{\bf Data and materials availability:} Data and code have been publicly released 
at\\
\url{https://ai-feynman.readthedocs.io}. 

\end{ack}

\section*{Broader Impact}

\subsection*{Who may benefit from this research}

Our research presumably has quite broad impact, since discovery of mathematical patterns in data is a central problem 
across the natural and social sciences. Given the ubiquity of {\it linear} regression in research, one might expect that there will significant benefits to a broad range of researchers also from more general symbolic regression once freely available algorithms get sufficiently good. 

\subsection*{Who may be put at disadvantage from this research}

Although it is possible that some numerical modelers could get their jobs automated away by symbolic regression, we suspect that the main effect of our method, and future tools building on it, will instead be that these people will simply discover better models than today.

\subsection*{Risk of bias, failure and other negative outcomes}

Pareto-optimal symbolic regression can be viewed as an extreme form of lossy data compression that uncovers the simplest possible model for any given accuracy. To the extent that overfitting can exacerbate bias, such model compression is expected to help. Moreover, since our method produces closed-form mathematical formulas that have excellent interpretability compared to black-box neural networks, they make it easier for humans to interpret the computation and pass judgement on whether it embodies unacceptable bias. This interpretability also reduces failure risk. 

Another risk is automation bias, whereby people overly trust a formula from symbolic regression when they extrapolate it into an untested domain. This could be exacerbated if symbolic regression promotes scientific laziness and enfeeblement, where researchers fit phenomenological models instead of doing the work of building models based on first principles. Symbolic regression should inform but not replace 
traditional scientific discovery.

Although the choice of basis functions biases the discoverable function class, our method is 
agnostic to basis functions as long as they are mostly differentiable. 

The greatest potential risk associated with this work does not stem from it failing but from it succeeding:
accelerated progress in symbolic regression, modularity discovery and its parent discipline, program synthesis, could hasten 
the arrival of artificial general intelligence, which some authors have argued humanity still lacks the tools 
to manage safely \cite{russell2019human}.
On the other hand, our work may help accelerate research on intelligible intelligence more broadly, and 
powerful future artificial intelligence is probably safer if we understand aspects of how it works than if it is an inscrutable black box.

\clearpage
\bibliographystyle{unsrt}
\bibliography{feynman2arxiv_v2}

\clearpage
\appendix

{\bf\Large Supplementary material}
\bigskip

Below we provide additional technical details about how we implement our method and numerical experiments.

\section{Testing for generalized symmetry}

\begin{figure}[h]
\centerline{
\includegraphics[width=\linewidth]{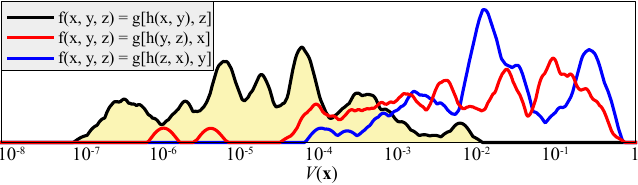}
}
\caption{Distribution of $V(\x_i)$ for the function from Figure~2, revealing that evidence for
the generalized symmetry $f(x,y,z)=g[h(x,y),z]$ (shaded distribution) is stronger than for 
$f(x,y,z)=g[h(x,z),y]$ (blue curve) or $f(x,y,z)=g[h(y,z),x]$ (red curve). The curves are shown slightly smoothed for clarity.
\label{VhistogramFig}
}
\end{figure}

We showed that generalized symmetry can be revealed by 
$\vhat(\x',\x'')$  being independent of $\x''$. We will now describe how we test
for such $\x''$-independence numerically.
Given
a point $\x_i\in\mathbb{R}^k$ from our data set, 
we compute a set of normalized gradients $\vhat_j\equiv\vhat_i(\x'_i,\x''_j)$, where $\x''_j\in\mathbb{R}^{n-k}$ correspond to a sample of $m$ other data points,
and quantify the variation between then by the quantity
\beq{VdefEq}
V(\x)\equiv 1-\max_{|\mu|=1} {1\over m}\sum_{j=1}^m(\muhat\cdot\vhat_j)^2
= 1-\max_{|\mu|=1} \muhat^t\V\muhat,\quad\hbox{where}\quad
\V\equiv {1\over m}\sum_{j=1}^m\vhat_j\vhat_j^t.
\eeq
We can intuitively interpret the optimal $\muhat$ as maximally aligned with the vectors $\vhat_j$ up to a sign.
\Eq{VdefEq} implies that our variation measure $V$ is simply one minus the smallest eigenvalue of $\V$, so 
$V$ ranges from $0$ when all $\vhat_j$ are identical to $1-{1\over m}$ when all eigenvalues are equal (equal to $1/m$, since $\tr\V=1$).
As illustrated in \fig{VhistogramFig}, we compute $V(\x_i)$ for each subset of up to $\ng$
input variables, and select the subset with the smallest median $V(\x_i)$ as the most promising generalized symmetry candidate. In our numerical experiments, we set the hyperparameter $\ng=3$ to save time, since we do not wish to consider all $2^n$ subsets for large $n$.

\section{Testing for generalized additivity}

We showed that generalized additivity holds when the function $s(x_1,x_2)$ from Equation~(3) from the main part of the paper is multiplicatively separable.
We will now describe how we test
for such separability numerically.
$s(x_1,x_2)$ being multiplicative separable is equivalent to $f(x_1,x_2)\equiv \ln{s(x_1, x_2)}$ being additively separable.
We numerically test the function $\ln{s_{NN}(x_1, x_2)}$ for additive separability using the normalized score $S$ defining
\beq{AddSepScore}
S[f] = \frac{|f_{,xy}|^2}{|f_{,xx} f_{,yy}| + |f_{,xy}|^2}.
\eeq
It is easy to see that $S[f]=0$ if $f$ is additively separable, and $S[f]>0$ otherwise.
If the median value of $S$ over all points $\x_i$ in the dataset is below a threshold $S_*$, 
we take this as evidence for generalized additivity and proceed as below. 
We found empirically that the threshold choise $S_*= 0.1$ produced robust results.
It is important to use smooth (not, \eg, ReLU) activation functions for this derivative-based test to be useful.

If this property holds, then we  recursively apply our 
algorithm to the two new 1-dimensional symbolic regression problems of discovering $a(x_1)$ and $b(x_2)$. If this succeeds and we are able to discover the functions 
$g(x_1)$ and $h(x_2)$ by symbolically integrating 
our solutions $g'=a$ and $h'=1/b$, then we have 
reduced the original problem to the same state as when we found compositionality above, now with $h(x_1,x_2)=g(x_1)+h(x_2)$.
Just as in that case, we simply replace the variables $\x$ in the original mystery data table by the single variable $h(\x)$ and recursively apply our AI Feynman algorithm to the new 1-dimensional symbolic regression problem of discovering how $f$ depends on $h$.

If we have determined that generalized additivity holds but the aforementioned method for discovering $g(x_1)+h(x_2)$ fails, 
we make a second attempt by training a neural network of the modular form $\fnn(x_1,x_2)=F[g(x_1)+h(x_2)]$ to fit the data. If this succeeds, 
we then recursively apply our AI Feynman algorithm to the three new 1-dimensional symbolic regression problems of discovering $F$, $g$ and $h$.

\section{Further details on success and failure modes}

Our paper reports which symbolic regression problems our method succeeded and failed on, as detailed in Tables~\ref{SuccessTable}-\ref{EqTableB}. Here we add specifics on how these successes and failures occurred.

\begin{table}[]
\begin{tabular}{|l|c|c|c|}
\hline
&Schmidt \& Lipson 2009&Udrescu \& Tegmark 2020&This paper\\
\hline
FDSR basic 100&71\%	&{\bf 100\%}	&{\bf 100\%}\\
FDSR harder 20&15\%	&85\%	&{\bf 90\%}\\
12 modular equations&42\%	&42\%	&{\bf 100\%}\\
10 probability distributions&60\%	&70\%	&{\bf 80\%}\\
\hline
\end{tabular}
\smallskip
\caption{\label{SuccessTable}
Fraction of symbolic regression problems solved for the benchmarks in the Feynman Database for Symbolic Regression (FDRS) and this paper.}
\end{table}

\paragraph{Success definition}

Given a data set $\{x_1,...,x_n,y\}$,
we use 90\% of the data to compute a Pareto-optimal set of candidate functions 
$\fhat_i(\x)$, then rank them based on their MEDL accuracy on the held-back 10\%
of the data. We count our method as successful only if the top-ranked function 
matches the true $f(\x)$ exactly, or, if the definition of $f$ involves irrational numerical
parameters, if these parameters are recovered to better than $0.01\%$ relative accuracy.

We considered an equation solved even if the top solution was not in the exact form presented in our
tables, but mathematically equivalent. For example, our method predicted that Equation~(12) in Table~4 was $w=\cos[\arccos(x)+\arccos(y)+\arccos(z)]$, which is
mathematically equivalent within the domain of our provided data set, where $x,y,z\in[-1,1]$.

For the problem of density estimation from samples, our goal was to obtain the correct normalized probability distributions.
The candidate functions on the Pareto-frontier were therefore
discarded unless they were non-negative and normalizable.
The surviving candidates were then normalized to integrate to unity by 
symbolic/numerical integration to obtain the appropriate normalization constant, and
quality-ranked by the surprisal loss function $$L_i =  -\sum \log\fhat_i(\x_k)$$
evaluated on the held-back test data.

\paragraph{Success examples}

Tables~\ref{EqTableA} and~\ref{EqTableB} below show the highest noise level allowing each of the 100 equations from the Feynman Database for Symbolic regression to be solved in the original paper analyzing it and in the present paper.

For many of the solved equations, the modularity discovery had to be used multiple times in order for the correct equation to be discovered, reflecting the power of the recursive algorithm. For example, for the quadruple velocity addition equation in Table 4, generalized symmetry was exploited twice. First, the code discovered that the first two velocities only enter in the combination $\frac{v_1+v_2}{1+v_1v_2}$, and these two variables were replaced by a new variable
$v_{12}$. The same method then discovered that $v_{12}$ and $v_3$ only enter in that same combination
$\frac{v_{12}+v_3}{1+v_{12} v_3}$, and thus the initial 3 variables $v_1$, $v_2$ and $v_3$ were replaced by a single variable $v_{123}$. Now the remaining equation had only 2 variables left, and was solved by brute force. 
In principle, this recursive method can be used to discover relativistic addition of an arbitrary number of velocities, by reducing the number of variables by one at each step. 

\paragraph{Failure examples}

Some of the most obvious failure modes we discussed in the conclusions of the main text.
Here we discuss some more subtle failure modes.
Firstly, it is worth noting that our definition of complexity is dependent on the chosen set of operations and does not always match our intuition.
For example, in fitting the probability distribution $$p(r, \theta) = \frac{1}{16} r^2 e^{-r} \cos^2\theta$$
of electron positions in the $n=2$, $l=1$, $m=0$ hydrogen orbital,
solutions with $\theta$-dependence $\cos\left(\cos\theta\right)$ are preferred over $\cos^2\theta$.
This is because, up to additive and multiplcative prefactors, the two formulas differ by at most approximately $2 \times 10^{-2}$ over our parameter range, but given a set of operations that includes only $\left\{\times, \cos\right\}$ denoted by "$*$" and ``$C$" respectively in reverse Polish notation, $\cos\left(\cos\theta)\right)$ (encoded as``$xCC$") is simpler than $\cos^2\theta$ (encoded as $``xCxC*"$).
In the presence of the imprecisions introduced by the normalizing flow, we were unable to perform the density estimation a level at which the accuracy for the correct $\cos^2\theta$ was preferred over the simpler alternative.

Furthermore, more interpretable approximations (e.g. Taylor expansions) are not always favored by our definition of complexity.
For example, in Figure 1, the unfamiliar solution
$$mc^2\left(\frac{1}{\cos{v/c}} - 1\right)$$
intermediate to the more familiar ${mv^2\over 2}$ and $mc^2\left(\frac{1}{\sqrt{1-v^2/c^2}} - 1\right)$ of  can be understood as a fourth-order expansion about $v=0$ of the exact formula.
Specifically, $mc^2\left(\frac{1}{\sqrt{1-v^2/c^2}} - 1\right) = \frac{mv^2}{2} + \frac{3mv^4}{8c^2} + O(v^6)$, and $mc^2\left(\frac{1}{\cos{v/c}} - 1\right) = \frac{mv^2}{2} + \frac{5mv^4}{24c^2} + O(v^6)$.
The Taylor expansions themselves are not preferred for reasons of complexity.

\begin{table*}[]
{\footnotesize
\begin{tabular}{|l|l|r|l|l|l|l|l|}
\hline
Feynman eq.  & Equation & Old Noise tolerance & New Noise tolerance\\
\hline                            

I.6.20a      & $f = e^{-\theta^2/2}/\sqrt{2\pi}$ & $10^{-2}$ & $10^{-1}$\\
I.6.20       & $f = e^{-\frac{\theta^2}{2\sigma^2}}/\sqrt{2\pi\sigma^2}$ & $10^{-4}$& $10^{-2}$\\
I.6.20b      & $f = e^{-\frac{(\theta-\theta_1)^2}{2\sigma^2}}/\sqrt{2\pi\sigma^2}$ & $10^{-4}$& $10^{-2}$ \\
I.8.14       & $d = \sqrt{(x_2-x_1)^2+(y_2-y_1)^2}$ & $10^{-4}$  &$10^{-1}$ \\
I.9.18       & $F = \frac{Gm_1m_2}{(x_2-x_1)^2+(y_2-y_1)^2+(z_2-z_1)^2}$ & $10^{-5}$       &   $10^{-3}$\\
I.10.7       & $m = \frac{m_0}{\sqrt{1-\frac{v^2}{c^2}}}$  & $10^{-4}$  & $10^{-2}$  \\
I.11.19      & $A = x_1y_1+x_2y_2+x_3y_3$ & $10^{-3}$  & $10^{-1}$  \\
I.12.1       & $F = \mu N_n$ & $10^{-3}$&  $10^{-1}$\\
I.12.1a      & $K = \frac{1}{2}m(v^2+u^2+w^2)$ & $10^{-4}$&  $10^{-1}$  \\
I.12.2       & $F = \frac{q_1q_2}{4\pi\epsilon r^2}$  & $10^{-2}$& $10^{-1}$\\
I.12.4       & $U = \frac{q_1}{4\pi\epsilon r^2}$ & $10^{-2}$ & $10^{-1}$ \\
I.12.5       & $F = q_2 E_f$ & $10^{-2}$ &$10^{-1}$  \\
I.12.11      & $F = q(E_f+B v \sin\theta)$ & $10^{-3}$ & $10^{-1}$ \\
I.13.12      & $U = Gm_1m_2(\frac{1}{r_2}-\frac{1}{r_1})$ & $10^{-4}$ &$10^{-1}$ \\
I.14.3       & $U = mgz$ & $10^{-2}$&$10^{-1}$  \\
I.14.4       & $U = \frac{k_{spring}x^2}{2}$  & $10^{-2}$&$10^{-1}$  \\
I.15.3x      & $x_1 = \frac{x-ut}{\sqrt{1-u^2/c^2}}$ & $10^{-3}$ & $10^{-3}$\\
I.15.3t      & $t_1 = \frac{t-ux/c^2}{\sqrt{1-u^2/c^2}}$ & $10^{-4}$ &  $10^{-3}$\\
I.15.1       & $p = \frac{m_0v}{\sqrt{1-v^2/c^2}}$ & $10^{-4}$ & $10^{-1}$\\
I.16.6       & $v_1 = \frac{u+v}{1+uv/c^2}$ & $10^{-3}$  &$10^{-2}$ \\
I.18.4       & $r = \frac{m_1r_1+m_2r_2}{m_1+m_2}$ & $10^{-2}$& $10^{-1}$ \\
I.18.12      & $\tau = rF\sin\theta$ & $10^{-3}$ & $10^{-1}$\\
I.18.14      & $L = mrv \sin\theta$   & $10^{-3}$ &$10^{-1}$ \\
I.24.6 & $E = \frac{1}{4} m (\omega^2+\omega_0^2) x^2$  & $10^{-4}$&$10^{-1}$ \\
I.25.13      & $V_e = \frac{q}{C}$  & $10^{-2}$& $10^{-1}$ \\
I.26.2       & $\theta_1 = \arcsin(n  \sin\theta_2)$ & $10^{-2}$ &$10^{-1}$ \\
I.27.6       & $f_f$    $ = \frac{1}{\frac{1}{d_1}+\frac{n}{d_2}}$    & $10^{-2}$& $10^{-1}$ \\
I.29.4       & $k = \frac{\omega}{c}$ &  $10^{-2}$ & $10^{-1}$\\
I.29.16      & $x = \sqrt{x_1^2+x_2^2-2x_1x_2\cos(\theta_1-\theta_2)}$  & $10^{-4}$ &$10^{-3}$  \\
I.30.3 & $I_* = I_{*_0}\frac{\sin(n\theta/2)}{\sin(\theta/2)}$  & $10^{-3}$ & $10^{-3}$\\
I.30.5       & $\theta = \arcsin(\frac{\lambda}{nd})$  & $10^{-3}$ & $10^{-1}$ \\
I.32.5       & $P = \frac{q^2a^2}{6\pi\epsilon_c^3}$     & $10^{-2}$&  $10^{-1}$\\
I.32.17 & $P = (\frac{1}{2}\epsilon c E_f^2)(8\pi r^2/3) (\omega^4/(\omega^2-\omega_0^2)^2)$     & $10^{-4}$&  $10^{-3}$ \\
I.34.8       & $\omega = \frac{qvB}{p}$  & $10^{-2}$ & $10^{-1}$\\
I.34.10       & $\omega = \frac{1+v/c}{1-v/c}\omega_0$ & $10^{-3}$&  $10^{-2}$\\
I.34.14      & $\omega = \frac{1+v/c}{\sqrt{1-v^2/c^2}}\omega_0$  & $10^{-3}$ & $10^{-3}$ \\
I.34.27      & $E = \hbar\omega$ & $10^{-2}$  & $10^{-1}$\\
I.37.4       & $I_* = I_1+I_2+2\sqrt{I_1I_2}\cos\delta$ & $10^{-3}$ & $10^{-2}$ \\
I.38.12      & $r = \frac{4\pi\epsilon\hbar^2}{mq^2}$   & $10^{-2}$& $10^{-1}$ \\
I.39.10       & $E = \frac{3}{2}p_F V$     & $10^{-2}$& $10^{-1}$ \\
I.39.11      & $E = \frac{1}{\gamma-1}p_F V$  & $10^{-3}$ & $10^{-1}$\\
I.39.22      & $P_F = \frac{n k_b T}{V}$        & $10^{-4}$ & $10^{-1}$\\
I.40.1       & $n = n_0e^{-\frac{mgx}{k_bT}}$     & $10^{-2}$ & $10^{-1}$\\
I.41.16      & $L_{rad} = \frac{\hbar\omega^3}{\pi^2c^2(e^{\frac{\hbar\omega}{k_bT}}-1)}$ & $10^{-5}$ & $10^{-4}$ \\
I.43.16      & $v = \frac{\mu_{drift}q V_e}{d}$   & $10^{-2}$ &$10^{-1}$ \\
I.43.31      & $D = \mu_e k_bT$                            & $10^{-2}$ & $10^{-1}$\\
I.43.43      & $\kappa = \frac{1}{\gamma-1}\frac{k_bv}{A}$  & $10^{-3}$ & $10^{-1}$ \\
I.44.4       & $E = n k_b T \ln(\frac{V_2}{V_1})$       & $10^{-3}$&  $10^{-1}$\\
I.47.23      & $c = \sqrt{\frac{\gamma pr}{\rho}}$   & $10^{-2}$& $10^{-1}$ \\
I.48.2       & $E = \frac{m c^2}{\sqrt{1-v^2/c^2}}$ & $10^{-5}$ & $10^{-3}$\\
I.50.26 & $x = x_1[\cos(\omega t)+\alpha\> cos(\omega t)^2]$     &  $10^{-2}$ &  $10^{-1}$  \\
\hline
\end{tabular}
\caption{Tested Equations, part 1
\label{EqTableA}
}
}
\end{table*}

\begin{table*}[]
\begin{tabular}{|l|l|r|l|r|l|l|l|}
\hline
Feynman eq.   & Equation  &Old Noise tolerance &New Noise tolerance\\
\hline       
II.2.42   & P     $ = \frac{\kappa(T_2-T_1)A}{d}$  & $10^{-3}$ &$10^{-1}$ \\
II.3.24   & $F_E = \frac{P}{4\pi r^2}$        & $10^{-2}$  & $10^{-1}$\\
II.4.23   & $V_e = \frac{q}{4\pi\epsilon r}$           & $10^{-2}$ & $10^{-1}$\\
II.6.11 & $V_e =\frac{1}{4\pi\epsilon}\frac{p_d\cos (\theta)}{r^2}$      & $10^{-3}$  & $10^{-1}$ \\
II.6.15a & $E_f = \frac{3}{4\pi\epsilon}\frac{p_d z}{r^5} \sqrt{x^2+y^2}$     & $10^{-3}$ & $10^{-2}$\\
II.6.15b & $E_f = \frac{3}{4\pi\epsilon}\frac{p_d}{r^3} \cos\theta\sin\theta$     & $10^{-2}$  & $10^{-2}$\\
II.8.7    & $E = \frac{3}{5}\frac{q^2}{4\pi\epsilon d}$   & $10^{-2}$ & $10^{-1}$\\
II.8.31   & $E_{den} = \frac{\epsilon E_f^2}{2}$          & $10^{-2}$ & $10^{-1}$\\
II.10.9   & $E_f = \frac{\sigma_{den}}{\epsilon}\frac{1}{1+\chi}$     & $10^{-2}$ &$10^{-1}$ \\
II.11.3 & $x = \frac{q E_f}{m(\omega_0^2-\omega^2)}$ & $10^{-3}$       & $10^{-2}$\\
II.11.7 & $n = n_0(1+ \frac{p_d E_f \cos\theta}{k_b T})$      & $10^{-2}$   & $10^{-1}$  \\
II.11.20  & $P_* = \frac{n_\rho p_d^2 E_f}{3 k_b T}$      & $10^{-3}$& $10^{-1}$ \\
II.11.27 & $P_* = \frac{n\alpha}{1-n\alpha/3}\epsilon E_f$ & $10^{-3}$   &  $10^{-1}$  \\
II.11.28  & $\theta = 1+\frac{n\alpha}{1-(n\alpha/3)}$    & $10^{-4}$  &$10^{-2}$ \\   
II.13.17  & $B = \frac{1}{4 \pi \epsilon c^2}\frac{2I}{r}$ & $10^{-2}$&$10^{-1}$  \\
II.13.23  & $\rho_c = \frac{\rho_{c_0}}{\sqrt{1-v^2/c^2}}$           & $10^{-4}$ & $10^{-2}$\\
II.13.24  & $j = \frac{\rho_{c_0}v}{\sqrt{1-v^2/c^2}}$     & $10^{-4}$ & $10^{-1}$\\
II.15.4   & $E = -\mu_M B \cos\theta$               & $10^{-3}$&  $10^{-1}$\\
II.15.5   & $E = -p_d E_f\cos\theta$  & $10^{-3}$ & $10^{-1}$\\
II.21.32  & $V_e = \frac{q}{4\pi\epsilon r(1-v/c)}$    & $10^{-3}$  &  $10^{-1}$\\
II.24.17 & $k = \sqrt{\frac{\omega^2}{c^2}-\frac{\pi^2}{d^2}}$       & $10^{-5}$ &  $10^{-2}$\\
II.27.16  & $F_E = \epsilon c E_f^2$        & $10^{-2}$& $10^{-1}$ \\
II.27.18  & $E_{den} = \epsilon E_f^2$            & $10^{-2}$& $10^{-1}$ \\
II.34.2a  & $I = \frac{qv}{2\pi r}$            & $10^{-2}$&  $10^{-1}$\\
II.34.2   & $\mu_M = \frac{q v r}{2}$                       & $10^{-2}$&  $10^{-1}$\\
II.34.11  & $\omega = \frac{g_{\_} q B}{2m}$         & $10^{-4}$ & $10^{-1}$\\
II.34.29a & $\mu_M = \frac{q h}{4\pi m}$       & $10^{-2}$&$10^{-1}$  \\
II.34.29b & $E = \frac{g_{\_} \mu_M B J_z}{\hbar}$ & $10^{-4}$&  $10^{-1}$\\
II.35.18 & $n = \frac{n_0}{\exp(\mu_m B/(k_b T))+\exp(-\mu_m B/(k_b T))}$      & $10^{-2}$& $10^{-2}$\\
II.35.21  & $M = n_\rho \mu_M \tanh(\frac{\mu_M B}{k_b T})$     & $10^{-4}$ &$10^{-4}$ \\
II.36.38 & $f = \frac{\mu_m B}{k_b T}+\frac{\mu_m\alpha M}{\epsilon c^2 k_b T}$      &  $10^{-2}$&  $10^{-1}$\\
II.37.1   & $E = \mu_M(1+\chi)B$   & $10^{-3}$  & $10^{-1}$\\
II.38.3   & $F = \frac{Y A x}{d}$         & $10^{-3}$&  $10^{-1}$\\
II.38.14  & $\mu_S = \frac{Y}{2(1+\sigma)}$           & $10^{-3}$  &$10^{-1}$ \\
III.4.32  & $n = \frac{1}{e^{\frac{\hbar\omega}{k_bT}}-1}$    & $10^{-3}$ & $10^{-2}$ \\
III.4.33  & $E = \frac{\hbar\omega}{e^{\frac{\hbar\omega}{k_b T}}-1}$  & $10^{-3}$   &  $10^{-3}$\\
III.7.38  & $\omega = \frac{2 \mu_M B}{\hbar}$   & $10^{-2}$  & $10^{-1}$ \\
III.8.54  & $p_{\gamma}$    $ = \sin(\frac{E t}{\hbar})^2$     & $10^{-3}$  &  $10^{-3}$ \\
III.9.52  & $p_{\gamma}$    $ = \frac{\frac{p_d E_f t}{\hbar}     \sin((\omega-\omega_0)t/2)^2}{((\omega-\omega_0)t/2)^2}$ & $10^{-3}$&   $10^{-1}$\\
III.10.19 & $E = \mu_M\sqrt{B_x^2+B_y^2+B_z^2}$    & $10^{-4}$ & $10^{-1}$\\
III.12.43 & $L = n\hbar$ & $10^{-3}$  &  $10^{-1}$\\
III.13.18 & $v = \frac{2 E d^2 k}{\hbar}$           & $10^{-4}$&  $10^{-1}$ \\
III.14.14 & $I = I_0 (e^{\frac{q V_e}{k_b T}}-1)$   & $10^{-3}$  &$10^{-1}$ \\
III.15.12 & $E = 2U(1-\cos(kd))$    & $10^{-4}$ &  $10^{-1}$ \\
III.15.14 & $m = \frac{\hbar^2}{2E d^2}$    & $10^{-2}$  &  $10^{-1}$\\
III.15.27 & $k = \frac{2\pi\alpha}{nd}$              & $10^{-3}$ &$10^{-1}$ \\
III.17.37 & $f = \beta(1+\alpha \cos\theta)$     & $10^{-3}$ &  $10^{-1}$\\
III.19.51 & $E = \frac{-mq^4}{2(4\pi\epsilon)^2\hbar^2}\frac{1}{n^2}$      & $10^{-5}$&   $10^{-2}$  \\
III.21.20 & $j = \frac{-\rho_{c_0} q A_{vec}}{m}$      &  $10^{-2}$   &  $10^{-1}$\\

\hline
\end{tabular}
\caption{Tested Equations, part 2. 
\label{EqTableB}
}
\end{table*}

\end{document}